\documentclass[10pt,twocolumn,letterpaper]{article}

\usepackage{cvpr}              % To produce the CAMERA-READY version
\definecolor{cvprblue}{rgb}{0.21,0.49,0.74}
\usepackage[pagebackref,breaklinks,colorlinks,allcolors=cvprblue]{hyperref}
\usepackage{multirow}
\usepackage{colortbl}
\usepackage{makecell}
\usepackage{amssymb}
\usepackage{amsmath}
\usepackage{xcolor}
\usepackage{bbding}
\usepackage{listings}
\usepackage{minted}
\usepackage{placeins}
\usepackage{float}
\usepackage{afterpage}
\usepackage[accsupp]{axessibility}  % Improves PDF readability for those with disabilities.

\def\paperID{5683} % *** Enter the Paper ID here
\def\confName{CVPR}
\def\confYear{2026}

\title{AutoTraces: Autoregressive Trajectory Forecasting via Multimodal Large Language Models}

\author{{Teng Wang\thanks{Corresponding author:~Teng Wang.}}\qquad{Yanting Lu}\qquad{Ruize Wang}\\
Southeast University\\
{\tt\small \{wangteng, luyanting, wangruize\}@seu.edu.cn}}
\begin{document}
\maketitle
\begin{abstract}
We present AutoTraces, an autoregressive vision-language-trajectory model for robot trajectory forecasting in humam-populated environments, which harnesses the inherent reasoning capabilities of large language models (LLMs) to model complex human behaviors. In contrast to prior works that rely solely on textual representations, our key innovation lies in a novel trajectory tokenization scheme, which represents waypoints with \texttt{$<$point$>$} tokens as categorical and positional markers while encoding waypoint numerical values as corresponding point embeddings, seamlessly integrated into the LLM’s space through a lightweight encoder-decoder architecture. This design preserves the LLM’s native autoregressive generation mechanism while extending it to physical coordinate spaces, facilitates modeling of long-term interactions in trajectory data. We further introduce an automated chain-of-thought (CoT) generation mechanism that leverages a multimodal LLM  to infer spatio-temporal relationships from visual observations and trajectory data, eliminating reliance on manual annotation. Through a two-stage training strategy, our AutoTraces achieves SOTA forecasting accuracy, particularly in long-horizon prediction, while exhibiting strong cross-scene generalization and supporting flexible-length forecasting.

% Codes and models will be available at: .

\end{abstract}    
\section{Introduction}
\label{sec:intro}
Forecasting socially compliant trajectories in human-populated environments remains a fundamental challenge for autonomous systems. As mobile robotic platforms are increasingly deployed across campuses, shopping malls, and other public areas, predicting trajectories that adhere to social norms while ensuring safety has become critically important. While deep reinforcement learning (DRL) has dominated prior motion planning research~\cite{Chen2017,Ciou2018,Samsani2021,Gonon2023,Zhu2023,Flogel2025,Kathuria2025}, its reliance on trial-and-error learning presents practical limitations for deployment. The recent emergence of large-scale human-teleoperated datasets~\cite{Scan2022,JRDB2023,MuSoHu2023,gostanford2019,shah2021rapid} has established ``Imitation Learning'' as a promising paradigm for socially compliant trajectory forecasting.
%Social navigation empowers autonomous robots to move through human-populated environments by adhering to social norms while maintaining safety and user comfort. With the expanding deployment of robots in public spaces like campus and shopping mall, the demand for socially-aware navigation has become increasingly significant. While Deep Reinforcement Learning (DRL) has dominated prior research~\cite{Chen2017, Ciou2018, Samsani2021, Gonon2023,Zhu2023,Flogel2025,Kathuria2025}, its reliance on trial-and-error learning presents significant practical limitations. The recent emergence of large-scale, real-world human-teleoperated social navigation datasets~\cite{Scan2022, JRDB2023, MuSoHu2023, gostanford2019, shah2021rapid} has established ``Imitation Learning from Expert Demonstrations'' as a promising alternative paradigm.

\begin{figure}[!t]
\centering
\includegraphics[width=0.99\linewidth]{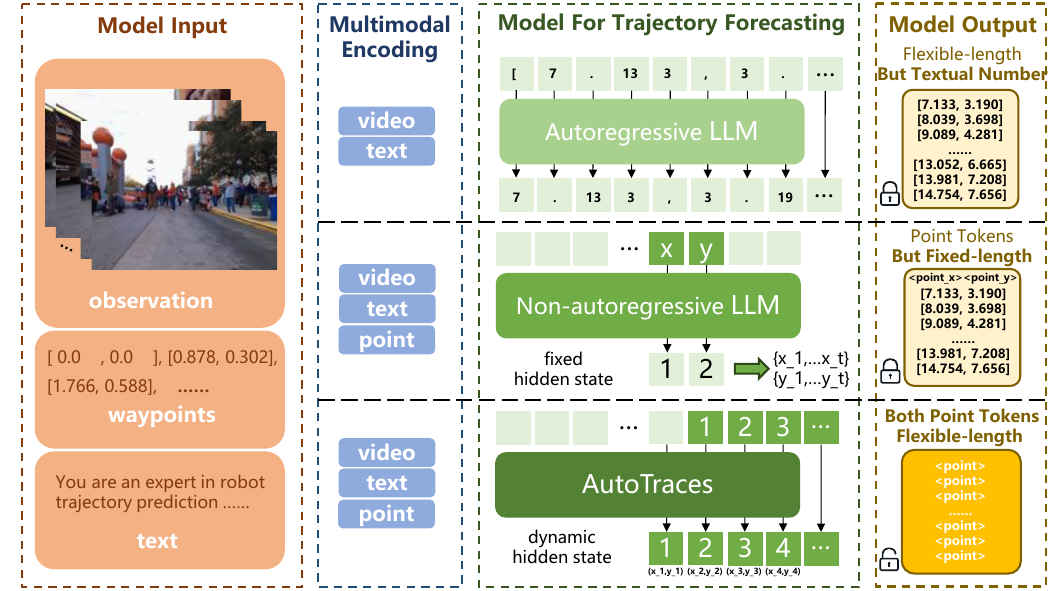}
\caption{Comparison between our AutoTraces and previous LLM solutions. Our method introduces point tokens and embeddings for waypoint representation, enabling autoregressive prediction.}
\label{fig1:problem}
\end{figure}

Given that visual perception serves as the primary sensory modality for most robotic platforms, the trajectory forecasting task is generally formulated as a multimodal planning problem that integrates historical trajectories with visual observations to generate target-conditioned future waypoints, thereby abstracting away low-level control.  Current state-of-the-art (SOTA) imitation learning approaches, including ViNT~\cite{ViNT2023}, NoMad~\cite{NoMaD2024}, and CityWalker~\cite{Citywalker2025}, typically employ learnable transformer decoders augmented by scene imagination~\cite{ViNT2023, Citywalker2025} or policy diffusion~\cite{NoMaD2024}, to predict fixed-length trajectory sequences. Despite considerable advances, these methods demonstrate limited generalization in open-world scenarios. We explain such deficiencies by the limited diversity and scale of expert demonstrations exacerbated by the end-to-end learning paradigm without human-like reasoning in between,  hindering genuine understanding of the underlying dynamics. 
%Based on these publicly available expert demonstrations, we aim to construct a comprehensive benchmark in this emerging field. Given that visual perception serves as the primary sensory modality for most robotic platforms, we formulate the task as multimodal planning that integrates historical trajectories with visual observations to generate future trajectory waypoints conditioned on targets, thereby abstracting away low-level control.  Current state-of-the-art (SOTA) imitation learning approaches, including ViNT~\cite{ViNT2023}, NoMad~\cite{NoMaD2024}, and CityWalker~\cite{Citywalker2025}, typically employ learnable transformer decoders augmented by scene imagination~\cite{ViNT2023, Citywalker2025} or policy diffusion~\cite{NoMaD2024}, to predict fixed-length trajectory sequences. Despite considerable advances, these methods demonstrate limited generalization across diverse social scenarios. We explain such deficiencies by the limited diversity and scale of expert demonstrations exacerbated by the end-to-end learning paradigm without human-like reasoning in between,  hindering genuine understanding of complex social interactions. 

Inspired by the success of large language models~(LLMs) in embodied intelligence~\cite{LM-Nav2023, Liu2023, NavGPT2024, ImagineNav2025, Kim24openvla}, we explore a trajectory forecasting framework that harnesses LLMs' inherent contextual reasoning capability to model complex human behaviors and generate socially compliant trajectories across diverse scenarios. Recent LLM-based methods~\cite{Kwon2024, Luo2025, Wang2024, Bae2024} for general trajectory prediction leverage LLMs' generative capabilities by framing prediction as QA tasks using \textit{textualized coordinates} (\cref{fig1:problem}), yet they suffer from token inefficiency and limited spatio-temporal modeling due to excessive coordinate tokenization. Parallel studies in broader spatio-temporal forecasting~\cite{UrbanGPT2024,Ma2025TPLLM} address this through task-specialized encoders to project structured data into LLM-compatible representations. However, these studies follow the \textit{non-autoregressive} paradigm, generating complete future sequences in a single pass from global representations obtained via static special tokens~\cite{UrbanGPT2024} or flattened hidden states~\cite{Ma2025TPLLM}, fundamentally limiting the modeling of temporal dynamics and flexible-length forecasting. Furthermore, leveraging visual observations to enhance LLMs' understanding of complex real-world human behaviors still remains an open challenge. 

\begin{figure*}[h]
\centering
\includegraphics[width=0.99\linewidth]{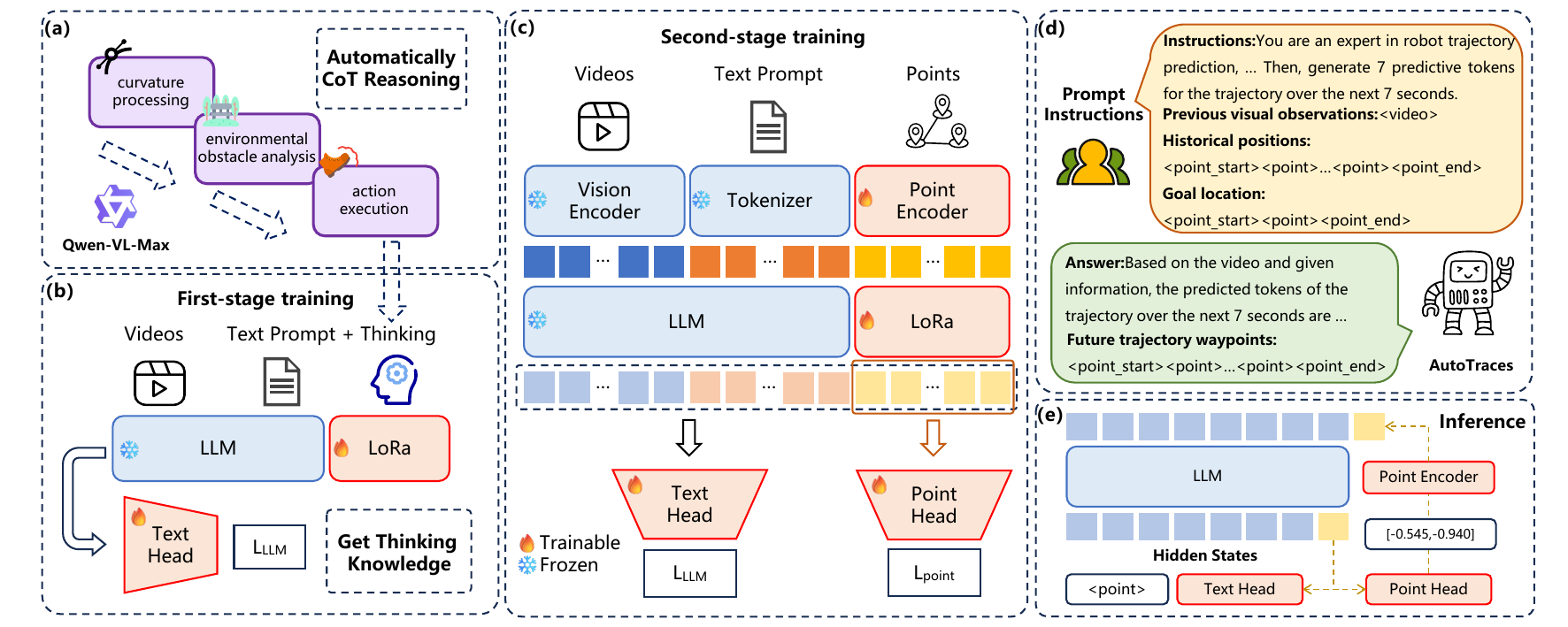}
\caption{The overall framework of our AutoTraces built upon LLaVa-Video~\cite{Zhang2024lLLaVaVideo} for socially compliant trajectory forecasting. In the first stage, the model is pre-trained on video-text pairs with reasoning prompts to acquire thinking knowledge. The second stage fine-tunes the model using trajectory waypoints (via a Point Encoder), aligning visual observations (via a Vision Encoder) and goal locations (via a Point Encoder). During inference, the model generates future trajectory waypoints as \texttt{$<$point$>$} tokens, enabling autoregressive predictions.}
\label{fig2:framework}
\end{figure*}

To tackle these challenges, we present \textbf{AutoTraces}, an \textit{autoregressive} vision-language-trajectory model that harnesses multimodal LLMs for socially compliant trajectory forecasting. The key innovation of our AutoTraces lies in a novel trajectory tokenization scheme that effectively bridges the gap between spatio-temporal trajectory patterns with LLMs' latent representations by introducing unique point embedding paradigm. Our model employs a unified \texttt{$<$point$>$} token to mark each waypoint--whether historical or future--individually and integrates waypoint values into corresponding point embeddings with textual and visual modalities seamlessly through a lightweight encoder-decoder framework (\cref{fig2:framework}). This design preserves LLMs' native autoregressive generation mechanism while extending it to physical coordinate spaces, facilitating long-term interaction modeling in trajectory data. Meanwhile, we introduce Chain-of-Thought (CoT) prompting to further enhance LLMs' understanding of complex social behaviors in video inputs. Rather than relying on manual annotation, we automate CoT generation by leveraging auxiliary LLM to perform structured reasoning about the underlying spatio-temporal relationships, facilitated by explicit geometric analysis of agent trajectories. Through our proposed two-stage training strategy, AutoTraces demonstrates superior prediction accuracy over existing SOTA methods and enhanced generalization across diverse scenes. Our contributions are summarized as follows. 
\begin{itemize}
    \item We propose a novel trajectory tokenization scheme that uses \texttt{$<$point$>$} tokens as categorical and positional markers, with waypoint values integrated into the LLM embeddings via an encoder-decoder architecture, enabling autoregressive generation of trajectories with enhanced spatio-temporal modeling. 
    \item We integrate CoT reasoning into dense trajectory prediction through an automated generation mechanism via multimodal LLMs, enhancing comprehension of complex social behaviors while eliminating manual annotation.
    \item Experimental results demonstrate our superiority in long-horizon trajectory prediction and cross-scene generalization, while supporting flexible-length forecasting.
\end{itemize}
%Departing from coordinate-as-text~\cite{liu2025reasonplan, Kwon2024, Luo2025, Wang2024, Bae2024}
% figure2

\section{Related Work}
\label{sec: Related Work}

%-------------------------------------------------------------------------
\noindent \textbf{Social Motion Planning.} Early methods primarily relied on handcrafted social rules and geometric constraints~\cite{SFM1995, Ferrer2013, Charalampous2016}, but struggled to model nuanced social behaviors in crowded scenarios. While DRL methods~\cite{Chen2017, Ciou2018, Samsani2021, Gonon2023,Zhu2023,Flogel2025,Kathuria2025} advanced the field, their trial-and-error learning paradigm faced deployment challenges. The shift to imitation learning enabled more flexible approaches, such as Panigrahi et al.~\cite{Panigrahi2023}, who fused RGB and LiDAR data through an MLP to predict trajectories. Subsequent transformer-based models--including ViNT~\cite{ViNT2023}, NoMad~\cite{NoMaD2024}, and CityWalker~\cite{Citywalker2025}--leveraged historical trajectories and visual context to infer fixed-length future trajectory. However, these methods exhibit limited generalization across diverse social scenarios. Recent LLM-based solutions such as Social-LLaVA~\cite{Payandeh2024} focus on generating high-level navigation commands through language-based reasoning. In comparison, our AutoTraces enables autoregressive prediction of waypoints through parameter-efficient adaptation of LLMs using trajectory tokenization scheme.

\noindent \textbf{LLMs for Spatio-Temporal Forecasting.}
A growing research direction explores spatio-temporal forecasting with LLMs, including applications to general trajectory prediction~\cite{Kwon2024, Wang2024, Chib2025, Luo2025, Bae2024} and traffic flow prediction~\cite{UrbanGPT2024, Ma2025TPLLM}. Among these approaches, one line of work~\cite{Kwon2024, Luo2025, Wang2024, Bae2024,hwang2025emma,Driess2023} leverages LLMs' generative capabilities by formulating spatio-temporal forecasting as question-answering tasks with text-based coordinate representations, while another category of methods~\cite{UrbanGPT2024, Ma2025TPLLM} utilizes LLMs' encoding capacity to process historical information and generate future predictions through introducing static special tokens~\cite{UrbanGPT2024} or output hidden state flattening~\cite{Ma2025TPLLM}. While LLMs have demonstrated strong performance in general time series understanding~\cite{LLMTime2023, UniTime2024, FPT2023, Time-LLM2024, AutoTimes2024}, socially compliant trajectories involve complex spatio-temporal patterns challenging LLMs’ generative ability. We bridge this gap by introducing a structured trajectory tokenization scheme, which preserves LLM’s native autoregressive generation mechanism while extending it to coordinate spaces.

\noindent \textbf{CoT Prompting for Robotics.} 
CoT prompting, originally developed to enhance LLM reasoning through step-by-step decomposition~\cite{Wei2022}, has recently been adapted to robotics for improved decision-making and task planning. By breaking down high-level instructions into sequential sub-goals, CoT prompting enables robots to perform multi-step reasoning, improving robustness in dynamic environments~\cite{Ahn2022, Zawalski2024,Ni2024, Driess2023, Brohan2023}. While CoT prompting has proven effective for high-level task decomposition and skill sequencing in robotics, its potential for dense trajectory prediction remains largely unexplored. To this end, we introduce automated CoT reasoning to trajectory generation by integrating geometric reasoning.
\section{Method}
\label{sec:method}

\subsection{Overall Framework}
We focus on socially compliant trajectory forecasting in human-populated environments by learning from a corpus of expert demonstrations. We formulate the task as a sequence generation task conditioned on multimodal inputs.  At each time step $t$, the agent receives a RGB observation $\mathbf{o}_t \in \mathbb{R}^{H\times W \times 3}$, its current position $\mathbf{x}_t \in \mathbb{R}^{2}$, and the goal location $\mathbf{g} \in \mathbb{R}^{2}$. The agent aims to generate a sequence of future trajectory waypoints $\mathbf{x}_{t+1: t+T} = \{\mathbf{x}_{t+1}, \cdots, \mathbf{x}_{t+2}, \mathbf{x}_{t+T}\}$ in Euclidean space conditioned on the previous visual observations $\mathbf{o}_{t-L:t} = \{\mathbf{o}_{t-L}, \cdots, \mathbf{o}_{t-1}, \mathbf{o}_{t}\}$, positional information $\mathbf{x}_{t-L:t} = \{\mathbf{x}_{t-L}, \cdots, \mathbf{x}_{t-1}, \mathbf{x}_{t}\}$ and the goal position $\mathbf{g}$. Typically, we take $L=8$ and $T$ varies to adapt to different navigation scenarios and robot configurations. 

As shown in~\cref{fig2:framework}, we present AutoTraces, an autoregressive vision-language-trajectory model built upon LLaVa-Video~\cite{Zhang2024lLLaVaVideo}. The model accepts a structured prompt that unifies historical waypoints and target goals into a shared \texttt{<point>} token space to ensure trajectory consistency, while incorporating video metadata (\textit{e.g.}, sequence length, frame rate) to enhance contextual reasoning. This formulation enables the model to interpret navigation contexts and generate trajectories of variable length as required. Furthermore, CoT reasoning is incorporated as a pre-training task to enhance AutoTraces's understanding of complex social interactions. The model is efficiently fine-tuned using Low-Rank Adaptation (LoRA) for both trajectory prediction and CoT reasoning tasks.%Furthermore, our LLM4SoNa jointly predicts future waypoints with chain-of-thought reasoning, revealing the spatio-temporal rationale for each navigational decision and improving interpretability. 

\subsection{Trajectory Tokenization}
\label{subsec:trajectory}

We propose an autoregressive multimodal LLM framework for flexible-length trajectory prediction. Departing from methods that directly output numerical coordinates as text~\cite{Kwon2024, Luo2025, Wang2024, Bae2024}, we introduce a structured tokenization scheme: a special \texttt{<point>} token is used to represent each 2D waypoint $\mathbf{x}_t \in \mathbb{R}^2$, while  \texttt{<point\_start>} and \texttt{<point\_end>} tokens demarcate trajectory boundaries. These tokens are integrated into the LLM model LLaVa-Video by extending the original vocabulary, effectively treating trajectory waypoints as a new output modality within the generative language framework.

To seamlessly integrate trajectory waypoints with textual and visual modalities, we design a lightweight encoder-decoder framework that operates directly in the LLM's native representation space. Specifically, we first introduce a \textit{Point Encoder} that represents waypoints using Transformer-style positional encoding, then feed them into MLPs, mapping physical coordinates into the LLM's latent space at each time step $t$, :
\begin{equation}
    \mathbf{e}_{t-i} = {\rm Point Encoder}(\mathbf{x}_{t-i}), \quad i = L, \dots, 0
\end{equation}
where $\mathbf{x}_{t-i} = (x_{t-i}, y_{t-i}) \in \mathbb{R}^2$ and $\mathbf{e}_{t-i} \in \mathbb{R}^D$ represents the $i$-th histroical waypoint and its encoded embedding, respectively. This mapping enables unified processing of trajectory, visual, and textual tokens in a shared feature space. The LLM takes as input a multimodal sequence comprising: (1) historical waypoint embeddings $\mathbf{E}_t = \{\mathbf{e}_{t-i}\}_{i=L}^{0}$, (2) visual embeddings $\mathbf{V}_t = \{\mathbf{v}_{t-i}\}_{i=L}^{0}$, where $\mathbf{v}_{t-i}$ encodes the visual observation $\mathbf{o}_{t-i}$, and (3) text prompt embeddings $\mathbf{P}_t$. Through autoregressive decoding, the model generates a sequence of $F$ future waypoint embeddings:
\begin{equation}
    \{\mathbf{\hat{e}}_{t+1}, \dots, \mathbf{\hat{e}}_{t+T}\} = {\rm LLM}(\mathbf{E}_{t}, \mathbf{V}_{t}, \mathbf{P}_{t}),
\end{equation}
where $\mathbf{\hat{e}}_{t+k}\in \mathbb{R}^D, k\in\{1,2,\cdots, F\}$, refers to the output feature embeddings of the $k$-th future waypoint. Finally, we decode the predicted waypoint embeddings back to physical coordinates via a \textit{Point Head}:
\begin{equation}
    \hat{\mathbf{x}}_{t+k} = {\rm Point Head}(\mathbf{\hat{e}}_{t+k}), \quad k = 1, \dots, T
\end{equation}
with $\hat{\mathbf{x}}_{t+k} = (\hat{x}_{t+k}, \hat{y}_{t+k}) \in \mathbb{R}^2$ denoting the coordinate of the predicted $k$-th future waypoint. Such a encoder-dencoder design enables the \texttt{<point>} modality to be naturally integrated into the LLM's token space without requiring architectural modifications to the base transformer. Our framework extends the LLM's autoregressive mechanism to physical coordinate, supporting structured waypoint prediction within a unified generative paradigm. Architectural details of point encoder/decoder are given in the \textbf{Appendix}. %This design preserves the original transformer architecture while extending it to process physical coordinates as a new token type. 

\iffalse
To align this new modality with existing textual and visual modalities, we design a \textit{Point Encoder} that maps physical waypoint coordinates from $\mathbb{R}^2$ to $\mathbb{R}^D$, embedding them into the model's shared latent space:
\begin{equation}
    \mathbf{e}_i = \text{PointEncoder}(\mathbf{x}_i), \quad i = t-L, \dots, t
\end{equation}
\noindent where $\mathbf{x}_i = (x_i, y_i) \in \mathbb{R}^2$ denotes the coordinates of the $i$-th waypoint, $\mathbf{e}_i \in \mathbb{R}^D$ denotes the corresponding embedding in the LLM's latent space, and $D$ denotes the feature dimension of the LLM's latent space. The encoded waypoint embeddings are processed autoregressively by the LLM:
\begin{equation}
    \{\mathbf{\hat{e}}_{t-L+1}, \dots, \mathbf{\hat{e}}_{t+1}\} = \text{LLM}(\{\mathbf{e}_{t-L}, \dots, \mathbf{e}_{t}\})
\end{equation}
\noindent where the model consumes $L$ historical waypoint embeddings to predict $F$ future waypoint embeddings, maintaining the same autoregressive generation mechanism as LLM.
Conversely, a \textit{Point Head} decodes the hidden states from $\mathbb{R}^D$ back to waypoint values in $\mathbb{R}^2$:
\begin{equation}
    \hat{\mathbf{x}}_i = \text{PointHead}(\mathbf{\hat{e}}_i), \quad i = t+1, \dots, t+F
\end{equation}
\noindent The \texttt{<point>} modality integrates naturally into the LLM's framework, enabling multi-modal feature fusion with text and vision in a shared token space, and supporting trajectory prediction without major architectural changes to the original LLM.
\fi

\subsection{Chain-of-Thought Reasoning with VLMs}
\label{subsec:cot}

To enhance the visual reasoning capability for complex behavioral understanding in video sequences, we introduce automatically generated Chain-of-Thought (CoT) reasoning as an intermediate representation. Rather than relying on manual annotation, we employ Qwen-VL-Max~\cite{bai2023qwen}, a powerful visual-language model, to produce CoT content directly from visual observations and trajectory data. Our approach not only eliminates the need for costly human labeling but also provides interpretable behavioral analysis that bridges visual perception and trajectory prediction. 

\begin{figure}[t!]
\centering
\includegraphics[width=0.99\linewidth]{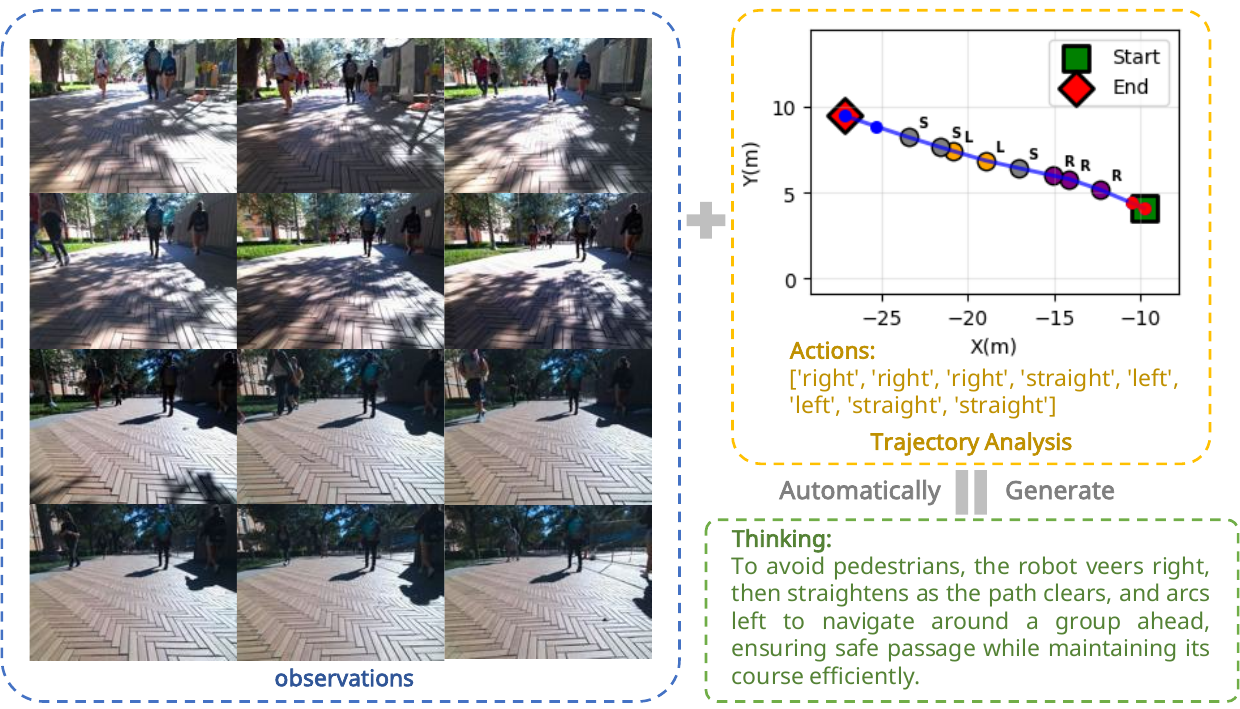}
\caption{Illustration of the CoT generation for AutoTraces, incorporating visual observations and trajectory analysis. Red points and lines denote the historical trajectory, while blue points and lines denote the ground-truth trajectory. Action annotations (R: right, L: left, S: straight) are marked along the trajectory.}
\label{fig3:prompt}
\end{figure}

Specifically, our CoT generation framework employs a structured prompting strategy that grounds reasoning in multimodal trajectory dynamics. We provide Qwen-VL-Max with synchronized inputs comprising ground-truth future trajectories and partial historical trajectories, along with their corresponding visual observations. In particular, we also incorporate explicit trajectory curvature analysis to facilitate Qwen-VL-Max for social reasoning, as illustrated in~\cref{fig3:prompt}. The curvature processing utilizes a sliding-window mechanism to decompose future trajectories into locally coherent segments, each categorized by directional patterns (e.g., ``straight," ``left," ``right"). This decomposition transforms continuous motion into a discrete meta-action sequence, whose symbolic representation structurally guides the reasoning generation while maintaining consistency between physical motion and linguistic rationale. The reasoning process follows a two-stage paradigm--environmental obstacle analysis followed by executable action derivation--ensuring each navigational decision remains visually grounded and logically traceable. 

%In the prompt of generating CoT reasoning, we incorporate both ground-truth trajectory points and visual observations as auxiliary information, along with curvature analysis of the trajectory. Using a sliding window approach, we segment the trajectory into local segments and classify their directional patterns (e.g., moving forward, turning left, or turning right), thereby generating a corresponding meta-action sequence that provides structural guidance for generating the reasoning chain. The overall reasoning process follows a two-step paradigm: first analyzing environmental obstacles, and then determining executable actions based on the analysis—mimicking human-like reasoning in navigation tasks. The model is trained to acquire the ability to capture the logical relationships between environmental obstacles and corresponding actions via fine-tuning on chain-of-thought data. Further implementation and training details will be elaborated in Sec.~\ref{subsec:training}

\begin{table*}[htbp]
\centering
\caption{Performance evaluation of trajectory prediction over 5 to 10 steps on SCAND dataset~\cite{Scan2022}, with an interval of 1s per step. Baselines with gray background are specifically trained and tested on fixed trajectory lengths. In contrast, ``Single Model'' variants are predominantly trained on T=10 sequences (except NoMad trained on T=8) and evaluated across all horizons under truncated-length settings.}
\label{tab:model_comparison}
\resizebox{0.75\linewidth}{!}{
\begin{tabular}{l c cc cc cc}
\toprule
\multirow{2}{*}{\textbf{Method}} & \multirow{2}{*}{\textbf{Single Model}} & \multicolumn{2}{c}{\textbf{T=5}} & \multicolumn{2}{c}{\textbf{T=8}} & \multicolumn{2}{c}{\textbf{T=10}} \\
\cmidrule(lr){3-4} \cmidrule(lr){5-6} \cmidrule(lr){7-8}
& & \textbf{L2~(m)}$\downarrow$ & \textbf{L1~(m)}$\downarrow$ & \textbf{L2~(m)}$\downarrow$ & \textbf{L1~(m)}$\downarrow$ & \textbf{L2~(m)}$\downarrow$ & \textbf{L1~(m)}$\downarrow$ \\
\midrule
\rowcolor{gray!10}GNM~\cite{Shah2023} & \textcolor{red}{\XSolidBrush} & \underline{0.855} & \underline{1.034} & 1.908 & 2.362 & 1.708 & 2.164 \\
GNM~\cite{Shah2023} & \textcolor{green}{\Checkmark}  & 0.895 & 1.122 & 1.456 & 1.826 & 1.708 & 2.164 \\
\rowcolor{gray!10}ViNT~\cite{ViNT2023} & \textcolor{red}{\XSolidBrush}  & 0.973 & 1.168 & 1.430 & 1.754 & 1.714 & 2.132 \\
ViNT~\cite{ViNT2023} & \textcolor{green}{\Checkmark}  & 0.908 & 1.106 & 1.435 & 1.762 & 1.714 & 2.132 \\
\rowcolor{gray!10}NoMad~\cite{NoMaD2024} & \textcolor{red}{\XSolidBrush}  & 1.167 & 1.440 & 1.625 & 2.102 & -- & -- \\
NoMad~\cite{NoMaD2024} & \textcolor{green}{\Checkmark}  & 1.060 & 1.318 & 1.625 & 2.102 & -- & -- \\
\rowcolor{gray!10}CityWalker~\cite{Citywalker2025} & \textcolor{red}{\XSolidBrush}  & 0.995 & 1.264 & 1.276 & 1.610 & 1.407 & 1.806 \\
CityWalker~\cite{Citywalker2025} & \textcolor{green}{\Checkmark}  & 0.862 & 1.096 & \underline{1.240} & \underline{1.572} & \underline{1.407} & \underline{1.806} \\
\hline
LLaVa-Video~\cite{Zhang2024lLLaVaVideo} & \textcolor{green}{\Checkmark} & 1.007 & 1.242 & 1.548 & 1.918 & 1.963 & 2.412 \\
\rowcolor{blue!10}\textbf{AutoTraces (ours)} & \textcolor{green}{\Checkmark}  & \textbf{0.674} & \textbf{0.856}  & \textbf{0.923} & \textbf{1.190}  & \textbf{1.089} & \textbf{1.384}  \\
\bottomrule
\end{tabular}}
\end{table*}

\subsection{Training and Inference}
\label{subsec:training}

\noindent\textbf{Autoregressive Prediction.}
Our AutoTraces fully leverage the autoregressive generation capability of LLMs to predict trajectories of theoretically arbitrary length. Rather than operating solely on \texttt{<point>} tokens, we retains the native textual and visual modalities (\cref{fig2:framework}~(d)), thereby preserving the model's pre-trained reasoning and task comprehension abilities to support trajectory prediction. Given a multimodal sequence of video, text, and historical \texttt{<point>} tokens, the objective for the model is to generate a subsequent sequence of future \texttt{<point>} tokens, the quantity of which must precisely match the requirement specified in the prompt instruction. As illustrated in~\cref{fig2:framework}~(e), upon generating a \texttt{<point>} token, it is routed to the \textit{Point Head} module, which decodes the corresponding hidden states into a waypoint coordinate. This coordinate is then processed by the \textit{point encoder} and appended to the input sequence. The updated sequence is fed back into the LLM to continue autoregressive generation until the model outputs the \texttt{</s>} (end-of-sequence) token, indicating the trajectory is complete. This autoregressive generation mechanism, where each newly predicted waypoint is immediately fed back to inform the next prediction, significantly enhances the model's capacity for long-horizon reasoning.

\noindent \textbf{Two-stage Training.} Our training pipeline employs a two-stage strategy designed to progressively instill visual reasoning capabilities while preserving the LLM's inherent knowledge. As illustrated in~\cref{fig2:framework}, the first stage focuses on learning interpretable reasoning patterns by taking historical observations $\mathbf{o}_{t-L:t}$ and past coordinates $\mathbf{x}_{t-L:t}$ as inputs to generate structured CoT rationales. This design explicitly decouples reasoning acquisition from trajectory prediction, allowing the model to leverage the LLM's pretrained knowledge while learning to ground visual and spatial contexts into coherent reasoning traces. At this stage, following standard LLM practice, we optimize the \textit{LoRA} layers and \textit{Text Head} using the standard cross-entropy loss, since model outputs are confined to text tokens. This CoT pre-training provides superior initialization for the LoRA weights over random initialization, thereby enhancing learning in the subsequent trajectory forecasting stage. In the second stage, we specializes the inherited model for trajectory forecasting by integrating the \texttt{<point>} modality into the native textual and visual modalities. To address the limitation of cross-entropy loss for regression--where it treats digits as discrete classifications and ignores error magnitude--we augment the objective with a trajectory point loss $\mathcal{L}_{\text{point}}$ to provide direct regression supervision on the output waypoints. The second-stage objective combines trajectory regression and sequence generation losses:
\begin{equation}
\begin{aligned}
\mathcal{L}_{\text{point}} &= \frac{1}{F} \sum_{i=t+1}^{t+F} \| \mathbf{x}_i - \hat{\mathbf{x}}_i \|_1, \\
\mathcal{L}_{\text{total}} &= \mathcal{L}_{\text{point}} + \mathcal{L}_{\text{LLM}},
\end{aligned}
\end{equation}
where $\mathcal{L}_{\text{LLM}}$ (cross-entropy) ensures the output sequence structure and length, while $\mathcal{L}_{\text{point}}$ ensures waypoint prediction accuracy. Parameter updates are confined to the \textit{LoRA} layers, \textit{Text Head}, \textit{Point Encoder} and \textit{Point Head}.
%\textcolor{blue}{enabling efficient knowledge transfer to the second stage without introducing premature bias from the \texttt{<point>} modality}%At \textbf{the second stage}, we introduce modules for the \texttt{<point>} token to specialize the inherited first-stage model in trajectory prediction. This specialization better aligns the model's pretrained visual reasoning capabilities with autoregressive forecasting. To address the limitation of cross-entropy loss in regression where it treats digits as discrete classifications and ignores error magnitude, we leverage our end-to-end pipeline to introduce a \textit{trajectory point loss} which directly supervises the numerical waypoints extracted from the model's output, enhancing regression performance.\:%\noindent\textbf{Prompt Design.} As illustrated in Fig.~\ref{fig3:prompt}, we encode both the historical positions and the goal location using a unified \texttt{<point>} modality. This ensures the alignment of input and output waypoint information, enabling the model to couple and comprehend information within the same token space. Furthermore, the instruction includes an interpretation of the video information, encompassing video length, frame rate, and a description of the task. This comprehensive prompt helps the LLM better understand the task content and generate a corresponding number of trajectory waypoints as required.
% \iffalse
% \begin{figure}[h]
% \centering
% \includegraphics[width=0.99\linewidth]{Fig/nexttoken.jpg}
% \caption{The process of the next-token prediction}
% \label{fig4:token}
% \end{figure}
% \fi
\section{Experiments}
\label{sec:Experiments}
%-------------------------------------------------------------------------
\begin{table*}[htbp]
\centering
\caption{Performance evaluation of cross-scene trajectory prediction on unseen GoStanford dataset~\cite{gostanford2019} for indoor and RECON dataset~\cite{shah2021rapid} for outdoor scenarios with prediction horizon ranging from 5 to 10 steps, with an interval of 1s per step.}
\label{tab:cross-scene}
\resizebox{0.75\linewidth}{!}{
\begin{tabular}{l l cc cc cc}
\toprule
\multirow{2}{*}{\textbf{Methods}} & \multirow{2}{*}{\textbf{Benchmark}} & \multicolumn{2}{c}{\textbf{T=5}} & \multicolumn{2}{c}{\textbf{T=8}} & \multicolumn{2}{c}{\textbf{T=10}} \\
\cmidrule(lr){3-4} \cmidrule(lr){5-6} \cmidrule(lr){7-8}  
&  & \textbf{L2~(m)}$\downarrow$ & \textbf{L1~(m)}$\downarrow$ & \textbf{L2~(m)}$\downarrow$ & \textbf{L1~(m)}$\downarrow$ & \textbf{L2~(m)}$\downarrow$ & \textbf{L1~(m)}$\downarrow$\\
\midrule
\multirow{2}{*}{\textbf{GNM}~\cite{Shah2023}}& GoStanford & 3.375 & 4.164 & 5.018 & 6.186 & 6.022 & 7.413  \\
& RECON & 1.754 & 2.328 & 3.141 & 4.168 & 4.117 & 5.448 \\
\midrule
\multirow{2}{*}{\textbf{ViNT}~\cite{ViNT2023}}& GoStanford & 2.980 & 3.688 & 4.320 & 5.348 & 5.192 & 6.378 \\
& RECON & 1.804 & 2.406 & 3.158 & 4.206 & 4.151 & 5.510  \\
\midrule
\multirow{2}{*}{\textbf{NoMad}~\cite{NoMaD2024}}& GoStanford & 3.470 & 4.252 & 5.108 & 6.224 & -- & --  \\
& RECON & 1.755 & 2.316 & 3.096 & 4.098 & -- & -- \\
\midrule
\multirow{2}{*}{\textbf{CityWalker}~\cite{Citywalker2025}}& GoStanford & 1.407 & 1.784 & 2.112 & 2.682 & 2.641 & 3.348 \\
& RECON & 1.955 & 2.500 & 3.082 & 3.938 & 3.877 & 4.966 \\
\midrule
\multirow{2}{*}{\textbf{LLaVa-Video}~\cite{Zhang2024lLLaVaVideo}}& GoStanford & 1.090 & 1.378 & 1.697 & 2.144 & 2.141 & 2.710  \\
& RECON & 1.943 & 2.472 & 3.289 & 4.182 & 4.211 & 5.358 \\
\midrule
\multirow{2}{*}{\textbf{AutoTraces (Ours)}}& GoStanford & 1.035 & 1.312 & 1.502 & 1.912 & 1.772 & 2.240  \\
& RECON & 1.547 & 1.974 & 2.302 & 2.936 & 2.837 & 3.614 \\
\bottomrule

\end{tabular}}
\end{table*}

\subsection{Experiment Settings}
\noindent\textbf{Datasets and Evaluation Metrics.} We evaluate our AutoTraces on the benchmark \texttt{SCAND} dataset~\cite{Scan2022}, a social navigation dataset capturing complex human-robot interactions in crowded public spaces. To assess cross-scene generalization, we further conduct experiments on two distinct real-world datasets: GoStanford~\cite{gostanford2019} for indoor and RECON~\cite{shah2021rapid} for outdoor scenarios. Following prior work~\cite{Citywalker2025}, we employ both L2 and L1 distances for evaluation. The L2-distance measures the average Euclidean displacement error between predicted and ground-truth trajectories, while the L1-distance provides robustness to outliers by computing the mean absolute error.

\noindent\textbf{Implementation Details.} 
All experiments are conducted on NVIDIA RTX 4090 GPUs. We construct our dataset from \texttt{SCAND}~\cite{Scan2022}, containing 24,512 training and 2,672 testing samples with strictly disjoint trajectories. We adopt the $160\times120$ image resolution from ViNT~\cite{ViNT2023}, and sample the data at 1 Hz. Each training sample uses a history of 8 steps and a future horizon randomly averaged between 5-10 steps, with a 1s interval per step. Our method is based on LLaVA-NeXT-Video-7B~\cite{Zhang2024lLLaVaVideo}, with its visual encoder CLIP~\cite{Radford2021LearningTV} and projector kept frozen. We employ a two-stage QLoRA~\cite{dettmers2023qlora} fine-tuning strategy. Stage 1: 2 epochs, batch size 4, learning rate $2\times10^{-5}$. Stage 2: 10 epochs, batch size 4, learning rate $2\times10^{-4}$. Both stages use AdamW-8bit~\cite{loshchilov2017decoupled,dettmers2022bit}, a cosine scheduler with 0.01 warmup, and LoRA rank $r=\alpha=32$.

\noindent \textbf{Baselines.} We consider several SOTA baseline methods for performance comparison, including GNM~\cite{Shah2023}, ViNT~\cite{ViNT2023}, NoMad~\cite{NoMaD2024}, CityWalker~\cite{Citywalker2025}, and LLaVa-Video~\cite{Zhang2024lLLaVaVideo}. The first four methods are specifically designed for robot trajectory prediction, while LLaVa-Video~\cite{Zhang2024lLLaVaVideo} exemplifies text-only adaptation of large vision-language models to trajectory forecasting, where continuous coordinates are serialized into discrete textual tokens for generative modeling. For non-autoregressive models (GNM, ViNT, NoMad, CityWalker), we consider two variants for fair comparison: (1) models specifically trained for individual horizon lengths, and (2) unified models trained exclusively with T=10 (except NoMad trained on T=8) and evaluated across all horizons under truncated-length settings. Unless otherwise specified, the unified models are used by default in the experiments. Please note that we exclude existing non-autoregressive LLM methods~\cite{UrbanGPT2024,Ma2025TPLLM} due to their incompatible task-specific encoders, but include a non-autoregressive variant in ablation studies to validate the advantage of autoregressive generation.

\subsection{Main Results}
We present a comprehensive evaluation of trajectory prediction performance across varying horizons (T=5, 8, 10) in~\cref{tab:model_comparison}. It could be noticed that our AutoTraces achieves SOTA performances across all evaluation metrics. For short-term prediction ($T=5$), AutoTraces attains L2 and L1 errors of 0.674m and 0.856m, respectively, surpassing the best-performing baseline GNM by 0.181m (L2) and 0.178m (L1). This advantage further amplifies in long-horizon settings. Specifically, AutoTraces achieves L2 and L1 errors of 1.089m and 1.384m at T=10 respectively, outperforming the second-performing CityWalker by 0.318m (L2) and 0.422m (L1). These results demonstrate the efficacy of our autoregressive approach in preserving prediction accuracy over extended time horizons.  Additionally, our AutoTraces demonstrates significant superiority over LLaVA-Video across all prediction horizons, validating the effectiveness of our proposed trajectory tokenization scheme combined with the CoT reasoning. %Furthermore, our approach substantially mitigates the progressive error accumulation commonly observed in non-autoregressive prediction methods. While baselines exhibit significant performance degradation as the horizon extends from T=5 to T=10--exemplified by CityWalker's L1 error increase from 1.096m to 1.806 (+0.71m)--our method demonstrates superior stability, with L1 error increasing only from 0.856m to 1.384m (+0.53m). This enhanced robustness against error accumulation underscores the fundamental advantage of autoregressive generation for long-term trajectory forecasting.

\begin{table*}[htbp]
\centering
\caption{Performance Comparison of Extended Trajectory Prediction on SCAND~\cite{Scan2022} with prediction horizon over 12-20 steps (1s/step).}
\label{tab:model_long-term}
\resizebox{0.80\linewidth}{!}{
\begin{tabular}{c cccc cccc}
\toprule
\multirow{2}{*}{\textbf{Horizon}} & \multicolumn{4}{c}{\textbf{LLaVa-Video}~\cite{Zhang2024lLLaVaVideo}} & \multicolumn{4}{c}{\textbf{AutoTraces (Ours)}} \\
\cmidrule(lr){2-5} \cmidrule(lr){6-9} 
 & \textbf{L2@S~(m)}$\downarrow$ & \textbf{L1@S~(m)}$\downarrow$ & \textbf{IEAcc} & \textbf{TPR} & \textbf{L2@S~(m)}$\downarrow$ & \textbf{L1@S~(m)}$\downarrow$ & \textbf{IEAcc} & \textbf{TPR} \\
\midrule
\textbf{12}  & 1.653 & 2.086 & \multirow{4}{*}{40.34\%} & \multirow{4}{*}{375.64} & 1.611 & 2.046  & \multirow{4}{*}{99.92\%} & \multirow{4}{*}{25.00} \\
\textbf{15}  & 2.057 & 2.597 & & & 1.857 & 2.354  & & \\
\textbf{18}  & 2.426 & 3.055 & & & 2.124 & 2.692  & & \\
\textbf{20}  & 2.584 & 3.244 & & & 2.324 & 2.950  & & \\
\bottomrule
\end{tabular}}
\end{table*}

\subsection{Cross-Scene Trajectory Prediction}
As shown in~\cref{tab:cross-scene}, we evaluate the generalization capability of all methods on two unseen datasets: GoStanford (indoor) and RECON (outdoor). The results demonstrate a clear advantage of autoregressive approaches--our AutoTraces and LLaVa-Video--over non-autoregressive baselines such as GNM, ViNT, NoMad, and CityWalker. This suggests that modeling trajectory prediction as an autoregressive decision-making process better captures temporal dependencies and leads to more robust performance in unseen scenes. Among autoregressive methods leveraging large foundation models, our AutoTraces consistently outperforms LLaVa-Video across all prediction horizons and benchmarks, with particularly notable gains observed in long-term prediction (T=10). Specifically, on the indoor GoStanford dataset, we achieve an L2 of 1.502 at T=8 and 1.772 at T=10, compared to LLaVa-Video’s 1.697 and 2.141, respectively. Similar trends are observed in outdoor RECON scenes, where our method reduces L2 error by 30.0\% at T=8 and 32.6\% at T=10. This significant superiority of AutoTraces over LLaVa-Video validate the effectiveness of our trajectory tokenization scheme and CoT reasoning in enabling more generalizable trajectory forecasting.
%As shown in~\cref{tab:cross-scene}, we evaluate the cross-scene generalization capability of all methods on two unseen datasets: GoStanford (indoor) and RECON (outdoor). While all methods experience performance degradation compared to their in-domain SCAND results (\cref{tab:model_comparison}), our LLM4SoNa demonstrates significantly better generalization. On the indoor GoStanford dataset, LLM4SoNa achieves L2 errors of 1.035m, 1.502m, and 1.772m for horizons T=5, 8, and 10 respectively, outperforming all baselines by a considerable margin. Notably, it reduces the L2 error by approximately 33\% compared to the strongest baseline CityWalker at T=10. Similar trends are observed on the outdoor RECON dataset, which covers more diverse scenarios including parking lots, suburban housing, sidewalks, and cafeterias. Here, LLM4SoNa achieves L2 errors of 1.547m (T=5), 2.302m (T=8), and 2.837m (T=10), consistently surpassing all competing methods. In particular, it reduces L2 and L1 errors by a margin of 1.040m and 1.352m compared to the strongest baseline CityWalker at T=10. The superior generalization performances demonstrate that LLM4SoNa learns fundamental navigation concepts instead of merely memorizing scene-specific patterns. This superiority could be attributed to our autoregressive trajectory generation framework, which unleashes the spatio-temporal reasoning capacity of LLMs by aligning trajectory waypoints with native textual and visual modalities via learnable special tokens while incorporating structured CoT reasoning. 

\subsection{Extending to Long-term Trajectory Prediction}
\label{subsec:Long-term Trajectory Prediction}
We further evaluate our model's generalization to long-term prediction horizons, as this flexible-length forecasting capability directly reflects its adaptability to diverse robotic platforms with varying velocities. We consider LLaVa-Video~\cite{Zhang2024lLLaVaVideo} for performance comparison. We report the \textbf{Instruction Execution Accuracy (IEAcc)} and \textbf{Tokens Per Response (TPR)}, combined with the average L2 and L1 errors under accurate execution (\textbf{L2@S} and \textbf{L1@S}) for each method. Specifically, IEAcc quantifies the model's capability to generate trajectories of the required length, while TPR measures the token count for a trajectory inference, which is hardware-independent and provides a consistent indicator of model efficiency. Our AutoTraces model was fine-tuned for only one epoch on a dataset comprising 1,336 samples (one-eighth of the total dataset), whereas LLaVA-Video was fine-tuned for one epoch on twice the amount of our data, each with a sequence length of 20. As shown in Table~\ref{tab:model_long-term}, our model achieves significantly higher instruction-following accuracy, attaining 99.92\% compared to 40.34\%, while achieving lower L1/L2 errors across all horizons under truncated-length settings. The instruction-following failures from LLaVa-Video primarily manifest as: (1) violation of length specifications, (2) failure to terminate generation, and (3) introduction of abnormal spatial dimensions (\textit{e.g.}, introducing a z-coordinate in (x,y) trajectories). The above superiority of our AutoTraces could be attributed to our point modality encoding mechanism, which facilitates the model's understanding of trajectory coordinates, enabling effective generalization across different scenarios with minimal fine-tuning samples and fine-tuning time, thereby significantly reducing computational costs. Consequently, for new domain adaptation, the entire language model requires no extensive retraining. Only the specialized prediction module and LoRA need lightweight fine-tuning, thereby leveraging the base model's pre-trained knowledge synergistically. Another key advantage of our point encoding mechanism is a dramatic reduction in TPR (25 vs. LLaVA-Video's 375). By representing each waypoint with a single token--unlike widely adopted text-based methods that require multiple tokens per coordinate--we provide an efficient pathway for numerical prediction in LLMs.

\subsection{Ablation Study}

\begin{figure}[h!]
\centering
\includegraphics[width=0.99\linewidth]{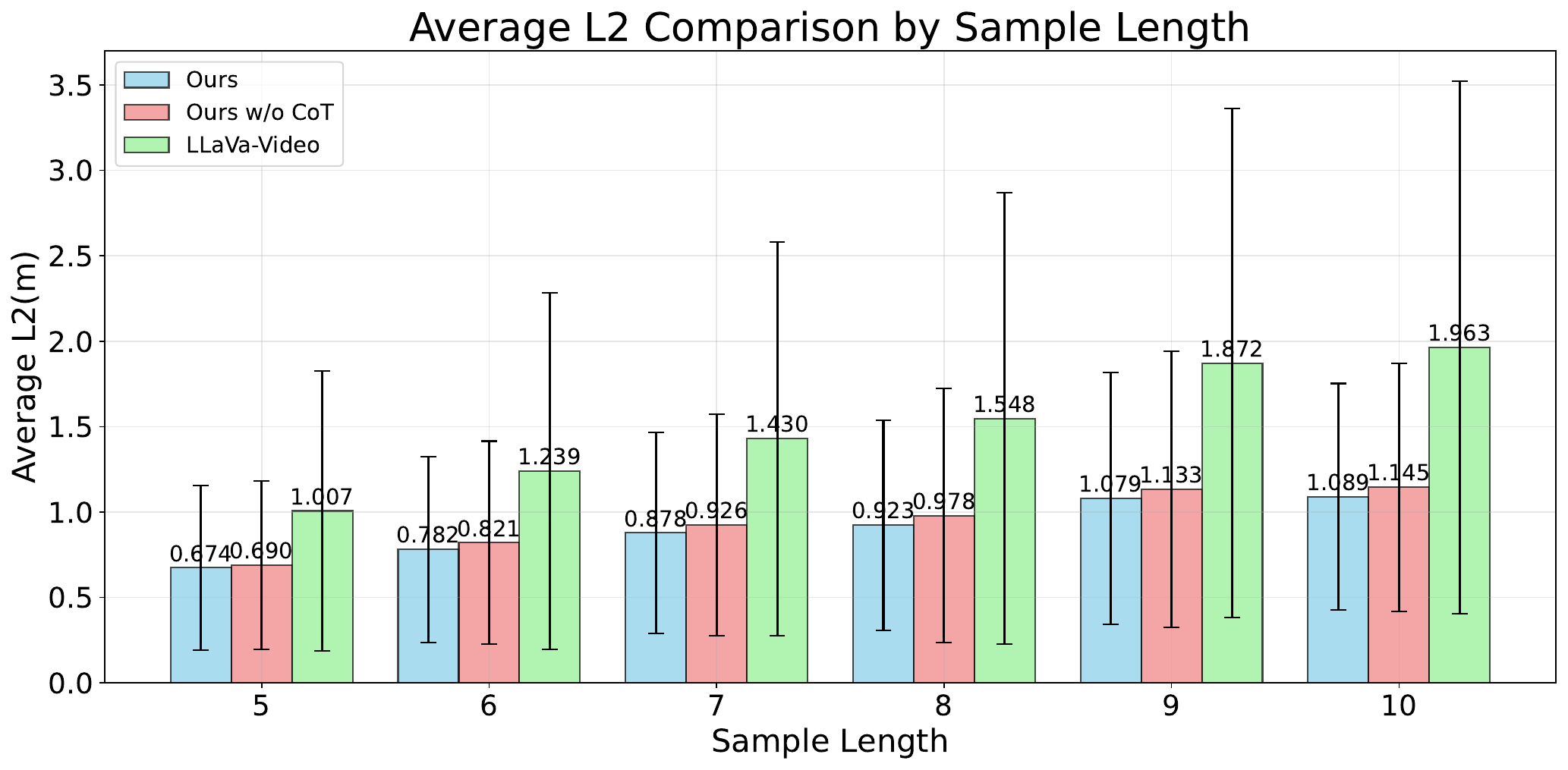}
\caption{Ablation study of our learnable waypoint tokenization scheme and CoT reasoning on \texttt{SCAND}~\cite{Scan2022} dataset.}
\label{fig4:ablation}
\end{figure}

\noindent\textbf{Effectiveness of Main Components.}
We conduct a comprehensive ablation study to evaluate the contributions of two key components in our AutoTraces: the waypoint embedding paradigm and the Chain-of-Thought (CoT) reasoning module. As shown in~\cref{fig4:ablation}, we compare three configurations on the SCAND~\cite{Scan2022} dataset: our full method (Ours), our method without CoT reasoning (Ours w/o CoT), and a baseline method LLaVa-Video that employs pure text representation for waypoints. The results demonstrate that our full method achieves superior performance across varying sample lengths, consistently maintaining the lowest L2 distance. Even without CoT reasoning, our method shows significant improvement over LLaVa-Video. For instance, at T=10, our method achieves an L2 distance of 1.145, substantially lower than LLaVa-Video's 1.963. This demonstrates the fundamental advantage of our learnable token representation over pure text encoding for waypoint modeling. In addition, the performance gap between our full model and the ablated version (Ours w/o CoT) clearly demonstrates the contribution of the CoT reasoning module. 

\begin{figure*}[h!]
\centering
\includegraphics[width=0.90\linewidth, height = 8.5cm]{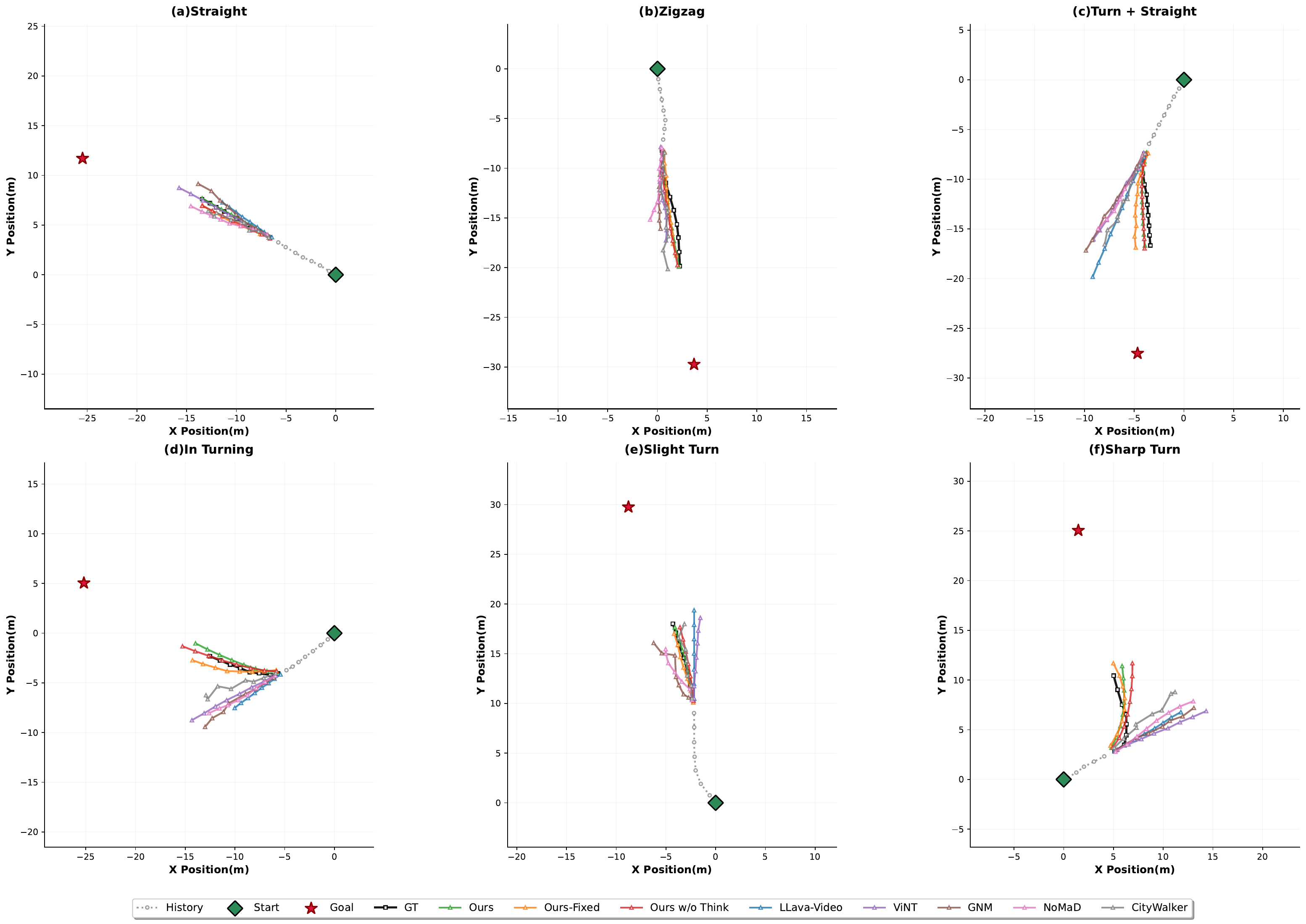}
\caption{Visualizations of predicted trajectory from our AutoTraces and other baselines on SCAND~\cite{Scan2022} dataset.}
\label{fig6:visualizations}
\end{figure*}

\begin{figure}[h!]
\centering
\includegraphics[width=0.99\linewidth]{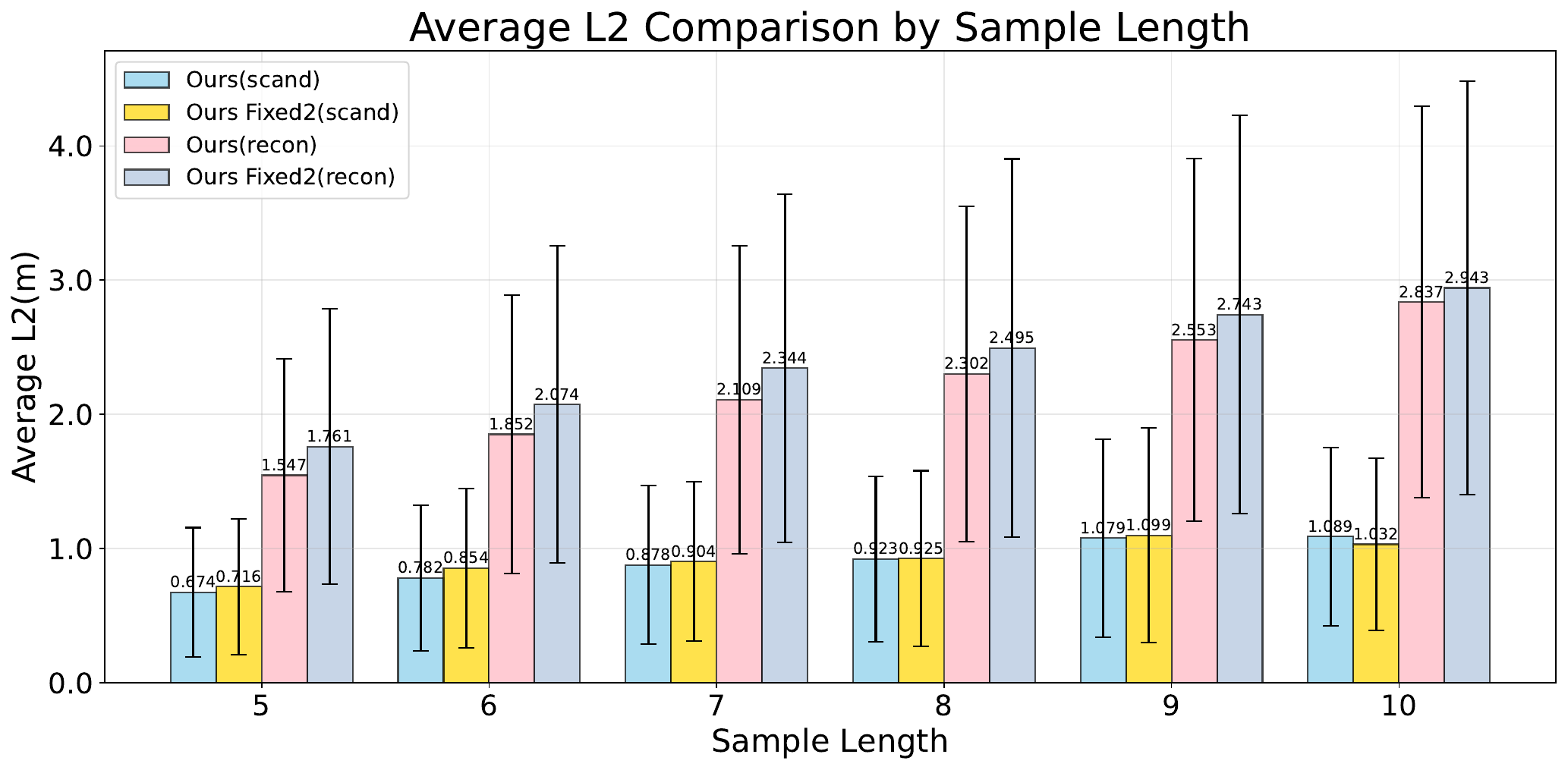}
\caption{Comparison between autoregressive and one-pass predictions on SCAND~\cite{Scan2022} and RECON~\cite{shah2021rapid} datasets.}
\label{fig5:ablation}
\end{figure}

\noindent\textbf{Autoregressive \textit{vs.} Single-pass Predictions.}  We conduct an ablation study on our model using a similar strategy as UrbanGPT~\cite{UrbanGPT2024} (see~\cref{fig1:problem}, middle) to illustrate the effectiveness of $\texttt{autoregressive}$ prediction. To be precise, we employ two fixed points to represent the x- and y-coordinates, and leverages fixed hidden states to decode trajectories of predetermined sequence lengths.
As shown in ~\cref{fig5:ablation}, on the SCAND dataset, our method achieves slightly better performance compared to the fixed model. Benefiting from the advantages of our point encoding strategy, the fixed model also achieves impressive performance.
In contrast, on the completely unseen RECON dataset, our autoregressive method demonstrates significantly stronger generalization, consistently outperforming the fixed-length baseline across all prediction horizons. This structural advantage stems from our point-wise autoregressive design, which ensures seamless alignment between historical and future waypoints during encoding and decoding, whereas the fixed method suffers from a encoder-decoder misalignment due to their decoupled design. This autoregressive forecasting mechanism with its inherent trajectory consistency enables better adaptation to dynamics variations (e.g., velocity), leading to stronger cross-domain generalization.  %This highlights the structural superiority of our method: the autoregressive generation is built upon point-wise encoding and decoding modules, which ensures consistency between historical and ground-truth trajectory waypoints during encoding and decoding. In the fixed method, however, the encoding of historical trajectories and the decoding of actual trajectories are decoupled, leading to a structural misalignment. Consequently, in unseen environments, our autoregressive approach—with its inherent trajectory consistency—better adapts to varying velocities and frequencies, substantially enhancing cross-domain generalization. %Due to the incompatibility of task-specific encoders in existing non-autoregressive LLM methods~\cite{UrbanGPT2024,Ma2025TPLLM}, direct comparisons in ~\cref{tab:model_comparison} are infeasible.

\subsection{Visualizations of Predicted Trajectory}
For an intuitive comparison, we visualize the predicted trajectories of different models in~\cref{fig6:visualizations}, which presents six representative navigation scenarios. In~\cref{fig6:visualizations}~(a), depicting smooth forward movement, all models correctly predict the direction, with our model achieving the highest accuracy. \cref{fig6:visualizations}~(b) illustrates a zigzag path typical of crowded environments, where none of the models adequately adjust their directions, resulting in unsatisfactory predictions. In the remaining subfigures~\cref{fig6:visualizations}~(c)–(f), we focus on the models’ performance in predicting turns of varying degrees. It can be observed that most non-LLM-based models fail to anticipate turning tendencies in most cases. In~\cref{fig6:visualizations}~(e), a few non-LLM-based models  perceive the turning intention in the slight turn scenario, only CityWalker accomplishes the turning maneuver, even though its trajectory accuracy remains suboptimal. In contrast, our LLM-based model effectively recognizes potential turning situations, leading to significantly improved trajectory prediction. Notably, LLaVA-Video, which relies solely on textual input, fails to produce reasonable turning behavior. This finding highlights the importance of the point modality in our model, which effectively guides trajectory prediction by integrating trajectory information with visual and textual information. Furthermore, the full model consistently achieves higher precision compared to its ablated versions across all scenarios.

\iffalse
\noindent\textbf{Data Flexibility:} Our model's data pipeline is compatible with arbitrary robot navigation datasets and supports variable-length inputs and outputs without any architectural modifications. This flexibility is achieved simply by adjusting the prompt text and the corresponding token sequence length. This capability facilitates seamless switching between long-term and short-term prediction horizons via zero-shot adaptation or fine-tuning, and provides significant advantages for cross-domain and cross-entity transfer learning. 
% Consequently, the model exhibits strong generalization power on completely unseen datasets.

\noindent\textbf{Model Extensibility:} Our framework is highly extensible due to its modular design. The \texttt{<point>} modality and its associated processing modules can be fully decoupled from the base LLM. This design effectively mitigates the issue of catastrophic forgetting in the LLM and allows for independent replacement of either the primary visual-language backbone or the point-processing components. This modularity offers substantial flexibility for adapting to different embodied tasks and integrating future state-of-the-art visual-language backbones.
\fi
\section{Conclusion}
\label{sec:Conclusion}
We introduced AutoTraces, an autoregressive vision-language-trajectory model built upon multimodal LLMs. Our approach integrates spatio-temporal trajectory data with LLM reasoning through a novel tokenization scheme that maps waypoints to special tokens and corresponding embeddings, preserving autoregressive generation while extending it to physical trajectory spaces. A lightweight encoder-decoder aligns trajectories with visual and textual inputs, and an automated CoT mechanism infers spatio-temporal relationships without manual annotation. Experiments show that AutoTraces achieves SOTA accuracy and remarkable cross-scene generalization, while supporting flexible-length predictions.

% \clearpage
\section*{Acknowledgment}
This work was supported by the National Natural Science Foundation of China (Grant No.62273093).
{
    \small
    \bibliographystyle{ieeenat_fullname}
    \bibliography{main}
}

% WARNING: do not forget to delete the supplementary pages from your submission 
%\input{sec/X_suppl}

\end{document}